\documentclass[conference,compsoc, onecolumn]{IEEEtran}

\usepackage{times}
\usepackage{latexsym}
\usepackage{booktabs}
\usepackage{graphicx} 
\usepackage{multirow}
\usepackage{url}
\usepackage{geometry}  % Add the geometry package
\makeatletter
\newcommand{\linebreakand}{%
  \end{@IEEEauthorhalign}
  \hfill\mbox{}\par
  \mbox{}\hfill\begin{@IEEEauthorhalign}
}
\makeatother

\ifCLASSOPTIONcompsoc
  \usepackage[nocompress]{cite}
\else
  \usepackage{cite}
\fi

\ifCLASSINFOpdf
\else
\fi
\begin{document}
\title{Data Augmentation to Improve Large Language Models in Food Hazard and Product Detection}

\author{
   \hspace{-1.1cm} \IEEEauthorblockN{Areeg Fahad Rasheed}
    \IEEEauthorblockA{
        \textit{College of Information Engineering} \\
        \textit{Al-Nahrain University}\\
        Baghdad, Iraq \\
        areeg.fahad@coie-nahrain.edu.iq
    }
    \and
    \IEEEauthorblockN{M. Zarkoosh}
    \IEEEauthorblockA{
        \textit{Independent Researcher} \\
        Baghdad, Iraq \\
        mzarkoosh@gmail.com
    }
    \linebreakand
    \IEEEauthorblockN{Shimam Amer Chasib}
    \IEEEauthorblockA{
        \textit{Department of Information Technology} \\
        \textit{Ministry of Labour and Social Affairs} \\
        Baghdad, Iraq \\
        shamamamir2017@gmail.com
    }
    \and
    \IEEEauthorblockN{Safa F. Abbas}
    \IEEEauthorblockA{
        \textit{Department of Cybersecurity Engineering} \\
        \textit{Al-Nahrain University} \\
        Baghdad, Iraq \\
        safaaf.abbas@gmail.com
    }
}

\maketitle

\begin{abstract}
The primary objective of this study is to demonstrate the impact of data augmentation using ChatGPT-4o-mini on food hazard and product analysis. The augmented data is generated using ChatGPT-4o-mini and subsequently used to train two large language models: RoBERTa-base and Flan-T5-base. The models are evaluated on test sets. The results indicate that using augmented data helped improve model performance across key metrics, including recall, F1 score, precision, and accuracy, compared to using only the provided dataset. The full code, including model training and the augmented dataset, can be found in this repository \url{https://github.com/AREEG94FAHAD/food-hazard-prdouct-cls}
\end{abstract}


% \fi
\IEEEpeerreviewmaketitle

\section{Introduction}
Food safety is a critical global concern, with millions of people affected by foodborne illnesses each year \cite{kanaan2023knowledge,zeki2024quality,uyttendaele2016food}. Rapid and accurate detection of food hazards is essential to prevent health risks and ensure consumer protection. However, the vast amount of textual data available in scientific literature, reports, and regulatory documents makes it challenging to efficiently classify and assess food-related risks\cite{van2018critical,european2011report}.

With the rapid advancement of Artificial Intelligence (AI) \cite{angelov2021explainable, wollowski2016survey}, particularly in the field of Natural Language Processing (NLP) a specialized subfield of AI dedicated to understanding, interpreting, and processing human language, we are now able to extract valuable insights from textual data with unprecedented efficiency \cite{rasheed2024taskcomplexity,rasheed2024exploring}. NLP has revolutionized automation across a wide range of applications, including text translation, grammar correction, information classification, text summarization, and question-answering \cite{rasheed2024mashee,rasheed2023arabic,zhao2023survey, sharma2018deep, wang2022research}.

In this paper, we will employ a machine learning technique called data augmentation \cite{Bayer_2022,adewumi2023t5} to enhance the performance of large language models (LLMs) for classifying hazard and product categories. By increasing the sample number of imbalanced classes within each category, we aim to address class imbalance. The augmented data will be generated using ChatGPT-4o-mini and subsequently used to fine-tune two LLMs: RoBERTa \cite{liu2019robertarobustlyoptimizedbert} and Flan-T5 \cite{chung2024scaling}.

The reset of the paper are organized as follows: In Section 2, we provide a brief overview of the models used in this study. Section 3 describes the dataset, including its characteristics before and after implementing data augmentation, as well as the proposed augmentation methodology. Section 4 details the fine-tuning process of the models and presents the main results obtained from training on both the original and augmented datasets. In Section 5, we will present a discussion of the findings. Finally, in the last section, we present the conclusions of the work.

\begin{figure*}[ht]
    \centering
    \includegraphics[width=0.99\linewidth]{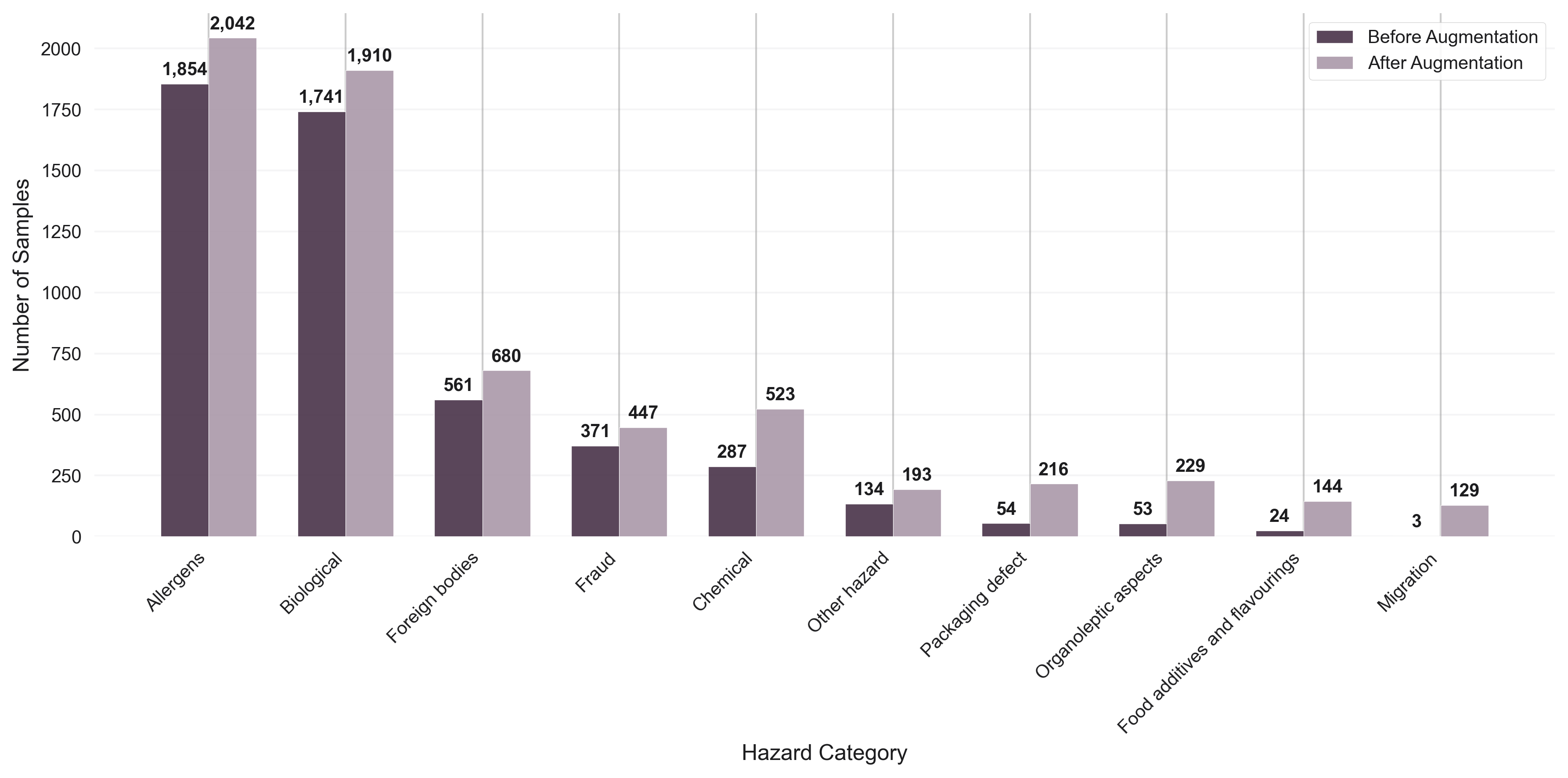}
    \caption{Distribution of Hazard Categories Before and After Data Augmentation}
    \label{fig:hazard-category_before}
\end{figure*}

\begin{figure*}[ht]
    \centering
    \includegraphics[width=0.99\linewidth]{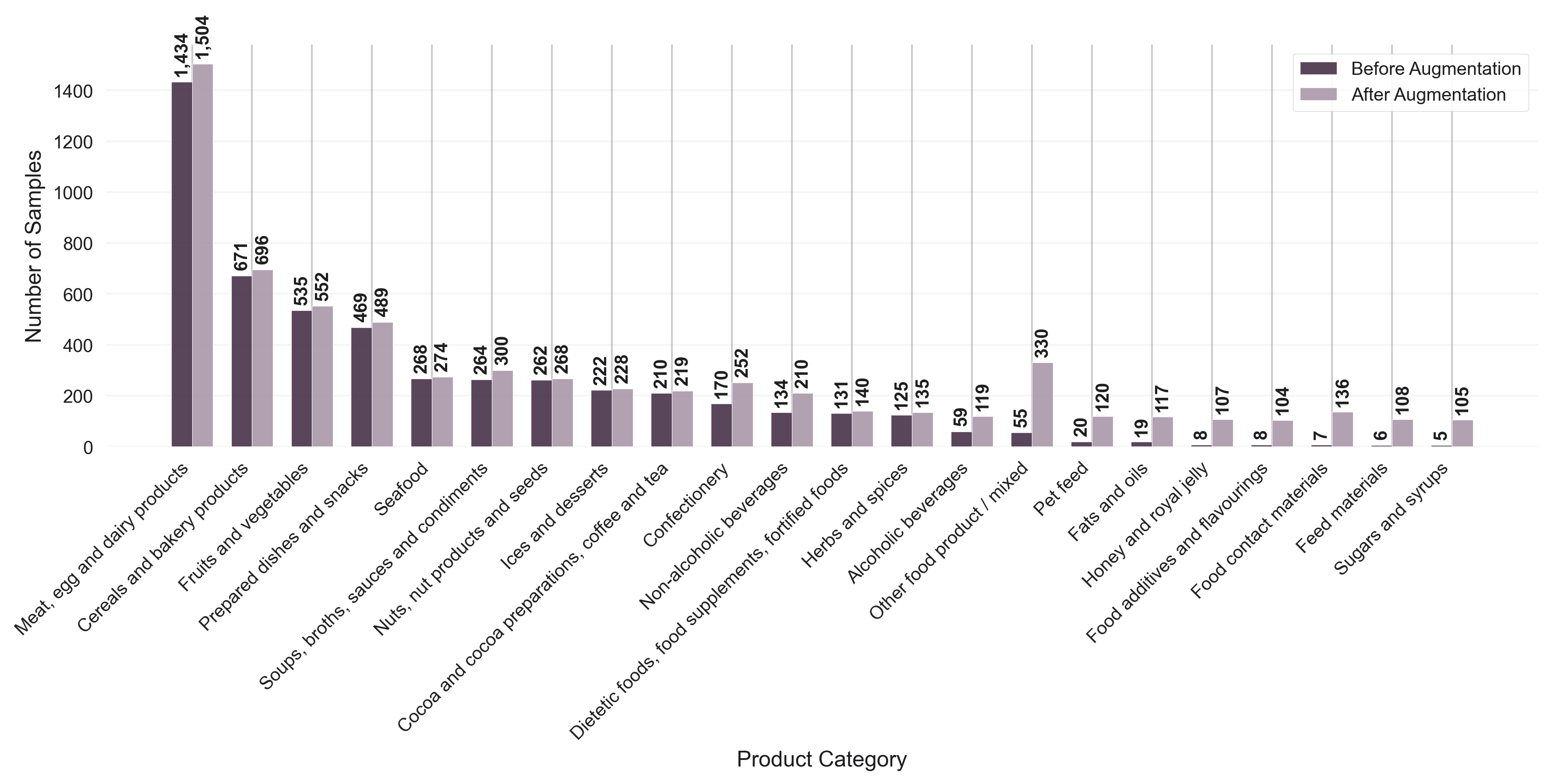}
    \caption{Distribution of Product Categories Before and After Data Augmentation}
    \label{fig:product-category_before}
\end{figure*}

\section{Large language models used}
In this work, we have used three large language models. The first one is ChatGPT-4o-mini for data augmentation. The second and third are RoBERTa and FLAN-T5 which are used for fine-tuning the dataset. The following section provides a brief description of each model.

\begin{itemize}

\item \textbf{ChatGPT-4o-mini:} ChatGPT-4o-mini is a version of ChatGPT released on May 14, 2024. It is a more efficient and resource-optimized model compared to previous versions. ChatGPT-40-mini is a multimodal model capable of processing text, images, and audio, making it suitable for a variety of tasks, including text generation, summarization, translation, and code assistance. It supports up to 128k tokens in a single context window. However, it is not publicly available as an API and can only be accessed through the ChatGPT-4o-mini interface \cite{aljanabi2023chatgpt, kalla2023study, briganti2024chatgpt}.

\item \textbf{RoBERTa} \cite{liu2019robertarobustlyoptimizedbert}: For fine-tuning the dataset for hazard and product category detection, we have used two models. The first one is RoBERTa, an LLM developed by Facebook AI Research that was released in 2019. It is an enhanced version of the BERT model. Like most LLMs, RoBERTa is used for many tasks such as topic categorization, summarization, etc. RoBERTa’s architecture is based on an encoder-only structure, meaning it does not have a separate decoder \cite{soliman2025comparative}. The maximum number of tokens supported by RoBERTa is 512 per input. There are many versions of RoBERTa; in this paper, we used the base version, which has 125 million parameters.

\item \textbf{Flan-T5} \cite{chung2024scaling}: The second model used is Flan-T5, a modified version of T5 that, like RoBERTa, is employed in various LLM applications. Since Flan-T5 is fine-tuned on a wide range of tasks, it is capable of zero-shot and few-shot learning. Instead of training the model extensively on a specific dataset, you can provide it with examples and instructions for tasks such as classification. Flan-T5 supports a full transformer architecture, meaning it has both encoder and decoder components \cite{vaswani2017attention}. Additionally, Flan-T5 supports a wider range of tokens, with the default being 512 tokens per input.
\end{itemize}

\begin{figure*}[ht]
    \centering
    \includegraphics[width=0.85\linewidth]{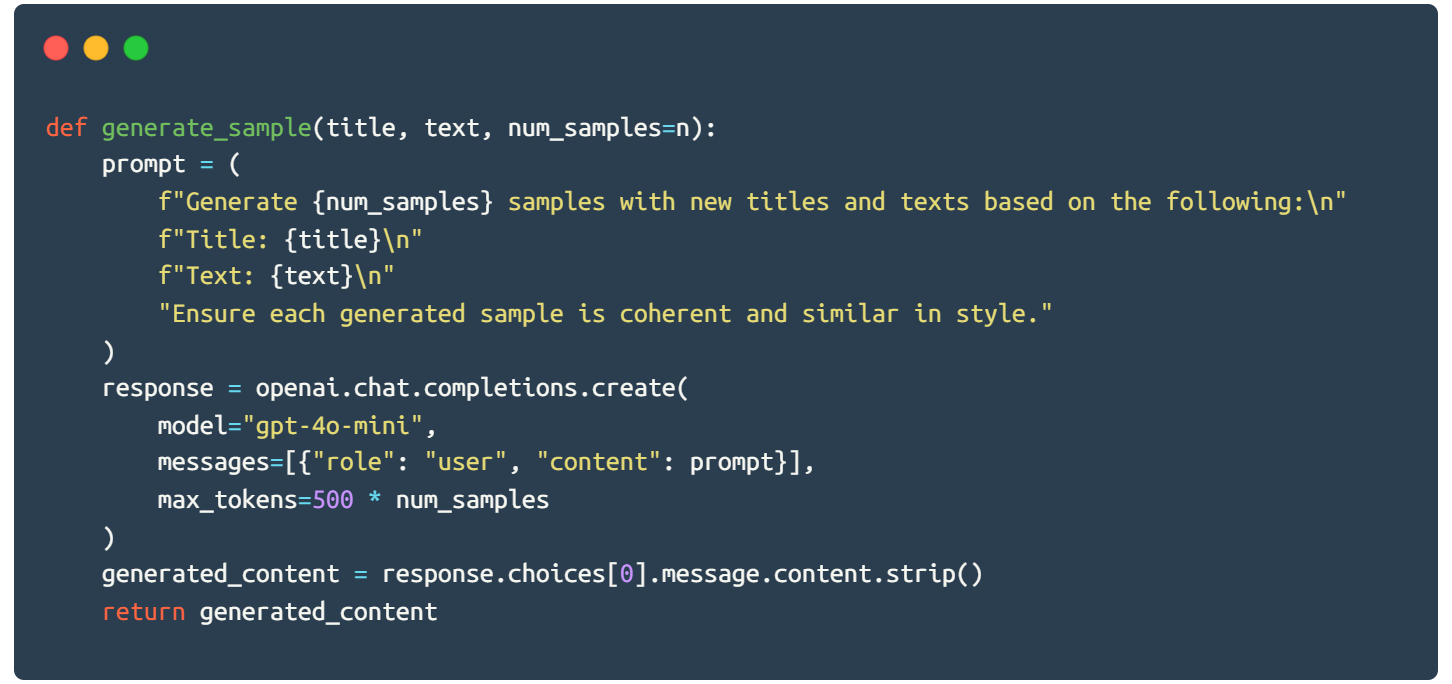}
    \caption{Prompt Used for Data Augmentation via ChatGPT-4o-mni}
    \label{fig:prompt}
\end{figure*}

\section{Dataset Pre- and Post-Augmentation}
The dataset for this paper consists of three files: the first is a training CSV file containing a total of 5,082 samples; the second is a development file with 565 samples; and the third is a test file with 997 samples. The following sections provide an in-depth analysis of the dataset's structure before and after augmentation was implemented \cite{FoodHazardSemEval2025}.

\subsection{Pre-Augmentation}
In Subtask 1, the goal is to classify the product and hazard category of each sample. Each sample contains multiple rows with attributes such as 'year', 'month', 'day', 'country', 'title', 'text', 'hazard-category', 'product-category', 'hazard', and 'product'. Since we are focusing only on Subtask 1, we will use the following attributes: 'title', 'text', 'hazard-category', and 'product-category'. The hazard category consists of ten classes, while the product-category includes 22 classes. The class names and the number of samples for each category are displayed in Figure \ref{fig:hazard-category_before} and Figure \ref{fig:product-category_before} respectively.

\subsection{Post-Augmentation}

For the augmentation process, we used the ChatGPT-4-omni version. We focused on increasing the number of samples for classes with a small sample number, such as the migration class. We utilized the ChatGPT API and sent a request with a prompt, as shown in Figure \ref{fig:prompt}.  After applying data augmentation, most hazard categories experienced an increase, particularly those with a smaller number of samples, as shown in Figure \ref{fig:hazard-category_before}. The number of migration samples increased from only 3 to 129, food additives and flavourings from 24 to 144, organoleptic aspects from 53 to 229, packaging defects from 54 to 216, and other hazards from 134 to 194. Additionally, chemical hazards increased from 287 to 523, fraud from 371 to 447, foreign bodies from 561 to 680, and biological hazards from 1,741 to 1,910. Meanwhile, the largest category, allergens, increased from 1,854 to 2,042. \ref{fig:hazard-category_before} show the statistic of hazard-category before and after augmentation.

Regarding the product category, which consists of 22 classes, we focused on increasing the number of samples in categories with smaller sample sizes. After applying data augmentation, the number of samples in the sugars and syrups category increased from 5 to 105, feed materials from 6 to 108, and food contact materials from 7 to 136. Similarly, the number of samples in honey and royal jelly increased from 8 to 107, food additives and flavourings from 8 to 104, and fats and oils from only 19 to 117. Pet feed increased from 20 to 120, other food products/mixed from 55 to 330, alcoholic beverages from 59 to 119, herbs and spices from 125 to 135, non-alcoholic beverages from 134 to 210, and confectionery from 170 to 252. Meanwhile, meat, egg, and dairy products increased from 1,434 to 1,504. The complete distribution of all categories, both before and after augmentation, can be found in Figure \ref{fig:product-category_before}.

After augmentation, the number of samples in the training set increased from 5082 to 6513.

\section{Fine-tuning Process}
In this section, we will show the impact of augmentation on the performance of two large language models, RoBERTa and FLAN-T5. The fine-tuning of both models has been done with specific parameters, as shown in Table \ref{tab:parameters_comparison}.

\begin{table*}[h]
  \centering
  \caption{Parameters of FLAN-T5-Base and RoBERTa Models.}
  \label{tab:parameters_comparison}
  \begin{tabular}{ccc}
    \toprule
    \textbf{Parameter} & \textbf{FLAN-T5-Base} & \textbf{RoBERTa}  \\
    \midrule
    Tokenizer         & T5Tokenizer                        & RobertaTokenizer                  \\
    Model            & T5ForConditionalGeneration        & RobertaForSequenceClassification  \\
    Pretrained Model & google/flan-t5-base               & roberta-base                      \\
    Max Length       & 256                                & 256                                \\
    Batch Size       & 8                                  & 16                                 \\
    Optimizer        & AdamW                              & AdamW                              \\
    Learning Rate    & 1e-4                               & 2e-5                               \\
    Epochs          & 100                                & 100                                 \\
    Input Format     & "Classify the hazard category: " + text & Text only                    \\
    Evaluation Metrics & Accuracy, F1, Precision, Recall & Accuracy, F1, Precision, Recall   \\
    Device           & CUDA                               & CUDA                               \\
    \bottomrule
  \end{tabular}
\end{table*}

\vspace{0.cm}

\begin{table*}[h]
  \centering
  \caption{Performance metrics before and after implementing augmentation for hazard-category and product-category classification}
  \label{tab:performance_metrics}
  \begin{tabular}{lllcccc}
    \toprule
    \textbf{Task} & \textbf{Model} & \textbf{Augmentation} & \textbf{Precision} & \textbf{Recall} & \textbf{Accuracy} & \textbf{F1-score} \\
    \midrule
    \multirow{4}{*}{Hazard category } 
      & \multirow{2}{*}{RoBERTa} 
        & Without       & 74.48 & 73.59 & 93.28 & 73.82 \\
        & & With augment  & 77.21 & 78.18 & 94.48 & 77.40 \\
      & \multirow{2}{*}{Flan-T5} 
        & Without       & 73.67 & 76.58 & 94.08 & 74.90 \\
        & & With augment  & 79.73 & 78.73 & 94.88 & 78.11 \\
    \midrule
    \multirow{4}{*}{Product category} 
      & \multirow{2}{*}{RoBERTa}
        & Without       & 72.16 & 74.32 & 79.84 & 71.82 \\
        & & With augment  & 77.45 & 76.27 & 80.24 & 76.26 \\
      & \multirow{2}{*}{Flan-T5}
        & Without       & 80.68 & 75.98 & 81.14 & 77.38 \\
        & & With augment  & 84.23 & 75.77 & 80.54 & 78.10 \\
    \bottomrule
  \end{tabular}
\end{table*}

Regarding the hazard category classification, the performance of RoBERTa and FLAN-T5 is similar, with FLAN-T5 achieving slightly better results, as shown in Table \ref{tab:performance_metrics} (73.82 and 74.90, respectively). After implementing augmentation, the performance of both systems improved. As shown in Table \ref{tab:performance_metrics}, both models saw an increase of approximately four points in F1-score. Regarding other performance metrics such as accuracy, recall, and precision, the results indicate that FLAN-T5 performed slightly better compared RoBERTa. However, both models showed improvements across all performance metrics when augmentation was applied to train the dataset. 

Regarding product category classification, the performance of FLAN-T5 is better than RoBERTa without augmentation. As shown in Table \ref{tab:performance_metrics}, the F1-score of RoBERTa is around 71, while FLAN-T5 achieves a higher score of 77. After implementing augmentation, the performance of both models improved. Notably, RoBERTa's F1-score increased from 71 to 76, while FLAN-T5 saw a smaller increase of about 1 percentage point.

\begin{figure}[ht]
    \centering
    \includegraphics[width=1\linewidth]{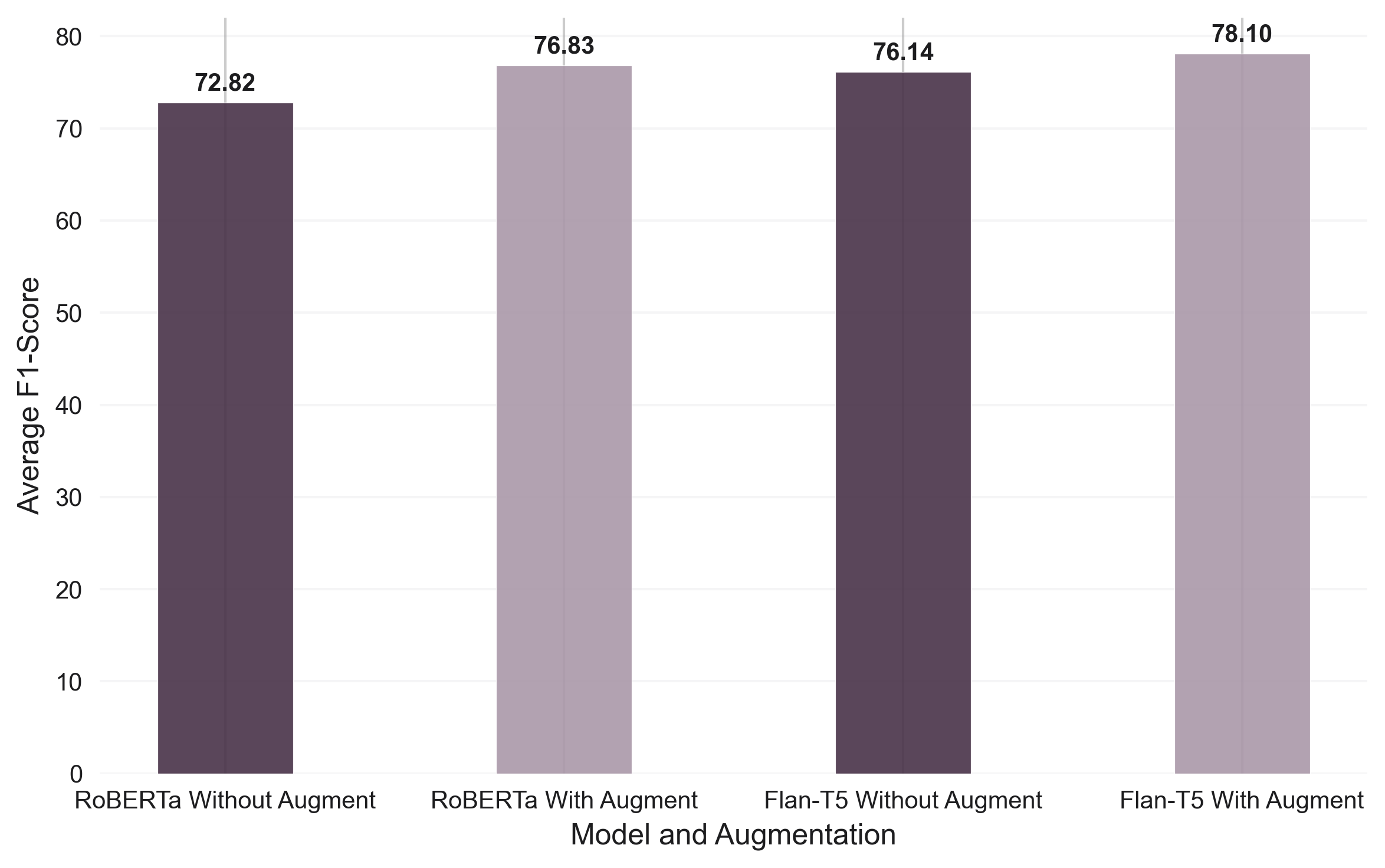}
    \caption{Average F1-score per model}
    \label{fig:averagef1}
\end{figure}

\begin{figure}[ht]
    \centering
    \includegraphics[width=1\linewidth]{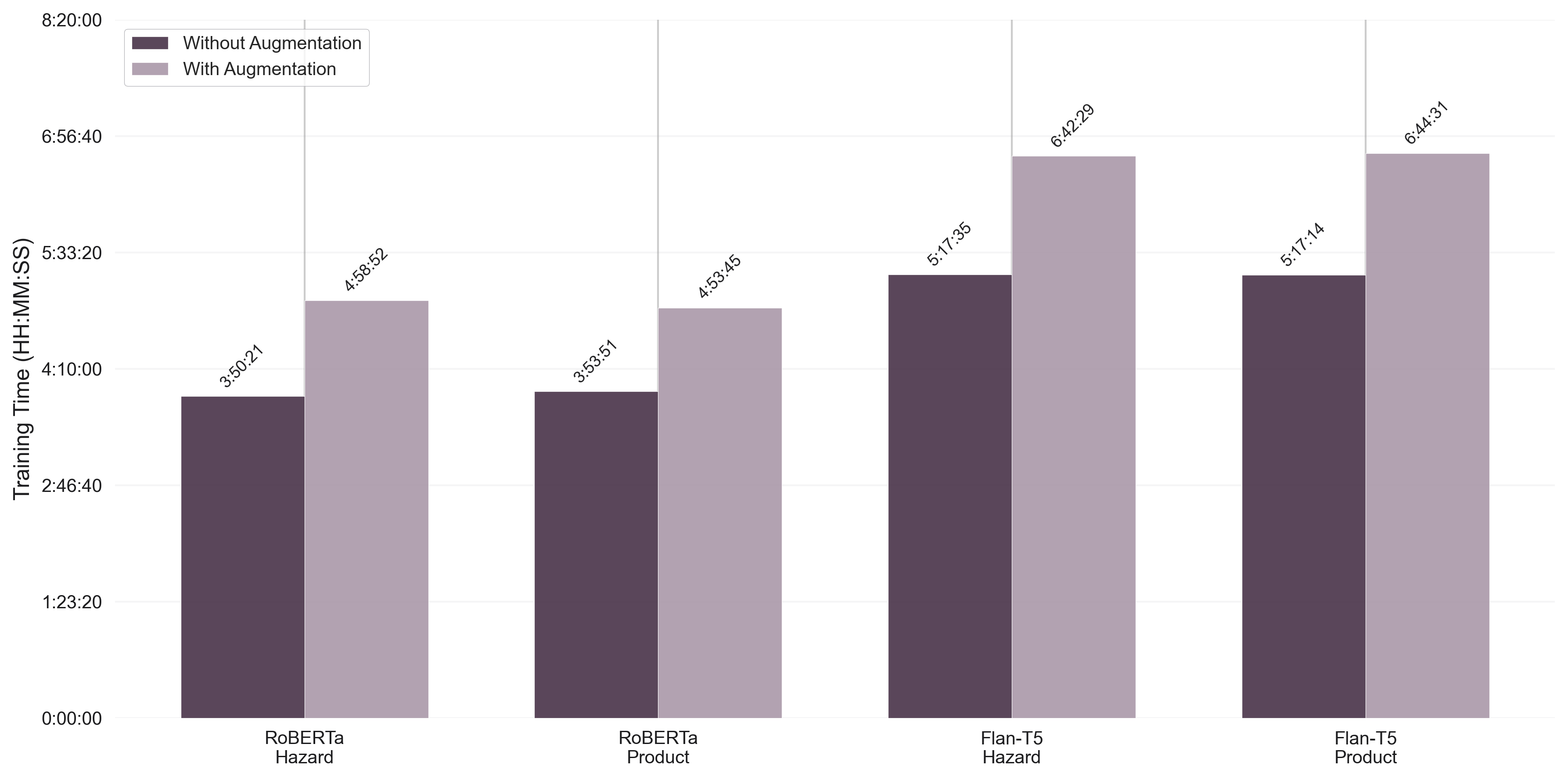}
    \caption{Model Training Times Comparison}
    \label{fig:training-time}
\end{figure}

With respect to training time, RoBERTa shows better performance compared to FlanT5 both with and without augmentation, as shown in Figure \ref{fig:training-time}. This is because RoBERTa's model size is smaller compared to FlanT5. Additionally, increasing the dataset size led to increased training time. Interestingly, the training time does not increase with higher numbers of classes even though the product-category classification task has more classes than the hazard-category task, the models trained in the same amount of time.

We also measured the average F1-score (macro) for hazard and product classification, Figure \ref{fig:averagef1} shows that Flan-T5 achieved the highest score of 78. RoBERTa with augmentation followed with 76.83, while models without augmentation performed lower.

\section{Discussion}
The results indicated three points, described below as the main reasons for the outcome.

\begin{itemize}
    \item [1-] Impace of model size: The highest performance was achieved using the FLAN-T5 model, which performed better with and without augmentation. The main reason is that FLAN-T5 has a larger model size than RoBERTa, with 248 million parameters compared to RoBERTa’s 125 million. An increased number of parameters enhances the understanding of complex text structures but also requires a longer training time compared to smaller models as shown in Figure \ref{fig:training-time}.
    \item [2-] Impact of Imbalanced Dataset: As shown in Figure \ref{fig:hazard-category_before} and \ref{fig:product-category_before}, the class migration category has only three samples, and the Syrups category has only five samples. This imbalance can increase the risk of overfitting, preventing the model from learning the structures of classes wih small number of samples effectively. After increasing the number of samples in these classes using augmentation, the model achieved better performance. 
    
    \item [3-] Impact of class number: After fine-tuning on the Hazard category, the model performed better both with and without augmentation compared to training on the Product category. The main reason for this is that the Product category has 22 classes, while the Hazard category only has 10 classes. The larger number of classes in the Product category increases the potential for misclassification, especially with under-fitting of small number of sample, compared to the more specialized Hazard category.
\end{itemize}

\section{Conclusion}

This work aims to investigate the impact of generated augmented data on Large Language Model (LLM) performance. In this study, we utilized two LLM models: RoBERTa and FlanT5. For data augmentation, we employed ChatGPT-4.0-mini version. The augmented dataset (original + augmented data) was used to train both models, and evaluation metrics were computed along with training time measurements.

The results indicate that across all performance metrics, FlanT5 demonstrates superior performance both with and without augmentation compared to RoBERTa. Data augmentation improved the performance of both models across all evaluation metrics. Training time was also measured, revealing that RoBERTa required less computational time compared to FlanT5, indicating higher computational efficiency.

Based on our findings, we provide two recommendations depending on available resources:
\begin{itemize}
\item  For balanced computational cost and performance, we recommend using RoBERTa with augmentation.
\item  When computational resources are not constrained, we recommend using FlanT5 with augmentation for optimal performance.

\end{itemize}

\bibliography{ref}
\bibliographystyle{IEEEtran}

\end{document}